\documentclass{article}
\usepackage{ijcai23}

\usepackage{times}
\usepackage{soul}
\usepackage{url}
\usepackage[hidelinks]{hyperref}
\usepackage[utf8]{inputenc}
\usepackage[small]{caption}
\usepackage{graphicx}
\usepackage{amsmath}
\usepackage{amsthm}
\usepackage{booktabs}
\usepackage{algorithm}
\usepackage{algorithmic}
\usepackage[switch]{lineno}

\usepackage{easyReview}

\usepackage{booktabs}
\usepackage{amsmath}
\usepackage{amssymb}
\usepackage{subfigure}
\usepackage{graphicx}
\usepackage{pifont}%
\usepackage{stfloats}

\title{Open-world Semi-supervised Novel Class Discovery}

\author{
Jiaming Liu\and
Yangqiming Wang\and
Tongze Zhang\and
Yulu Fan\and
Qinli Yang\And
Junming Shao\footnote{Corresponding author}\\
\affiliations
University of Electronic Science and Technology of China\\
\emails
\{liujiaming, leo\_wang, zhangtongze, ylfan\}@std.uestc.edu.cn,
\{qinli.yang, junmshao\}@uestc.edu.cn
}

\begin{document}

\maketitle

\begin{abstract}
Traditional semi-supervised learning tasks assume that both labeled and unlabeled data follow the same class distribution, but the realistic open-world scenarios are of more complexity with unknown novel classes mixed in the unlabeled set. Therefore, it is of great challenge to not only recognize samples from known classes but also discover the unknown number of novel classes within the unlabeled data. In this paper, we introduce a new \textit{Open-world Semi-supervised Novel Class Discovery} approach named OpenNCD, a progressive bi-level contrastive learning method over multiple prototypes. The proposed method is composed of two reciprocally enhanced parts. First, a bi-level contrastive learning method is introduced, which maintains the pair-wise similarity of the prototypes and the prototype group levels for better representation learning. Then, a reliable prototype similarity metric is proposed based on the common representing instances. Prototypes with high similarities will be grouped progressively for known class recognition and novel class discovery. Extensive experiments on three image datasets are conducted and the results show the effectiveness of the proposed method in open-world scenarios, especially with scarce known classes and labels.
\end{abstract}


\section{Introduction}

\begin{figure}[tp]
\centering 
\includegraphics[width = 0.40\textwidth]{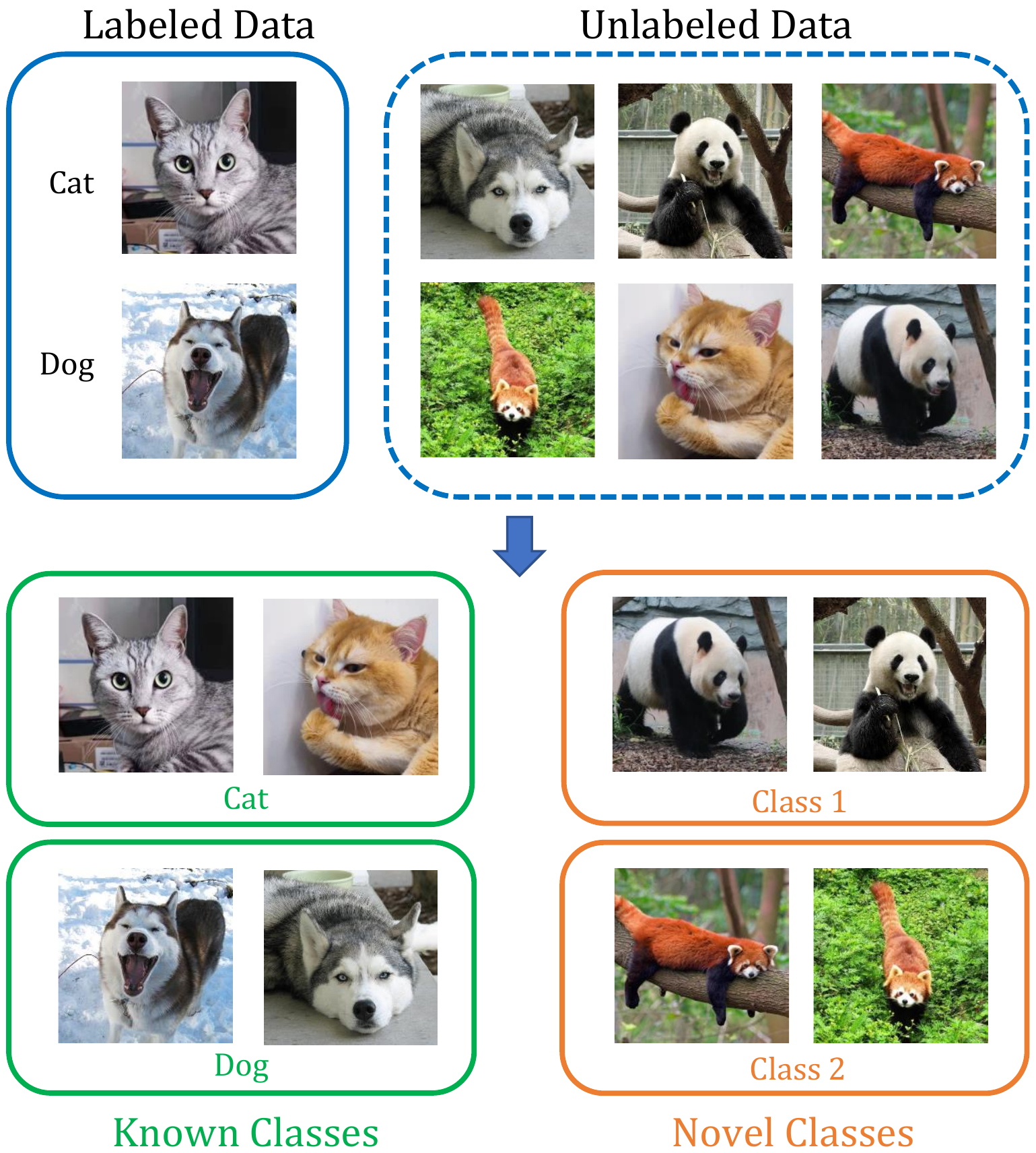}
\caption{
An illustration of the open-world semi-supervised novel class discovery task, where there exist some unknown novel classes in the unlabeled data in semi-supervised learning. The objective is to recognize the samples from known classes while simultaneously discovering the unknown number of novel classes within the unlabeled data.
}
\label{fig:seen-novel-pictures}
\end{figure}

Modern deep learning approaches have received widespread attention and progressed rapidly with labeled datasets. Despite the many strengths, most of the approaches are based on a close-world assumption, where the class distributions remain unchanged in the testing phase. But in the realistic open world, the close-world assumptions can barely hold and novel classes are likely to appear in the unlabeled set. For example, in the field of autopilot, the model needs to not only recognize the pre-trained traffic signs (known classes) but also discover the unknown obstacles (novel classes). For cybersecurity, managers need to detect and classify network intrusions as either existing types or unknown numbers of new types of attacks. Therefore, as shown in Figure \ref{fig:seen-novel-pictures}, compared to the close-world settings, the open-world scenarios are more complicated with the following challenges.

\textit{\textbf{Challenge 1}: How to recognize the known classes and detect the unknown class samples mixed in the unlabeled dataset?} 
Traditional semi-supervised learning tasks assume a close-world setting, where the classes in the unlabeled set are the same as the known labeled set.
But for open-world scenarios, unknown class samples are mixed in the unlabeled set. To tackle this problem, some robust semi-supervised learning methods \cite{guo2020DS3L,huang2021trash} are proposed, in which all detected novel samples are simply and crudely regarded as outliers. The recently proposed ORCA \cite{orca} can simultaneously cluster the different novel classes in the unlabeled set, but with the assumption that the number of novel classes is predefined.

\textit{\textbf{Challenge 2}: How to better cluster the novel classes in the unlabeled set with an extra disjoint labeled dataset?} 
To address this challenge, some novel class discovery methods \cite{han2019learning,han2019automatically} are proposed, which try to leverage useful knowledge from an extra labeled set to better cluster the novel classes in the unlabeled data. However, most of them hold the assumption that all samples are from the novel classes in the testing phase. Therefore, these methods do not have the ability to identify known class samples that are mixed with the unknowns in the test set. Recently, GCD \cite{gcd} generalizes the novel class discovery task to further recognize the known classes in the unlabeled set. But the predictions can only be made in a transductive way, which requires the whole test set.

\textit{\textbf{Challenge 3}: How to estimate the number of unknown classes and match the suitable cluster hierarchy?}
Some traditional methods \cite{pelleg2000x-means,hamerly2003G-means} and recent deep learning methods \cite{leiber2021dip,ronen2022deepdpm} are proposed to estimate the number of clusters in unsupervised clustering tasks. However, it is difficult to determine the clustering level since several different cluster hierarchies often exist in complex datasets. For example, the CIFAR-100 dataset has 100 classes and they can also be grouped into 20 superclasses. For the open-world setting, some labeled samples can be used to better determine the most suitable cluster hierarchy.

In this paper, we aim to address the aforementioned three open-world challenges for the recognition of known classes and the discovery of the arbitrary number of novel classes in the unlabeled data. To this end, a progressive bi-level contrastive learning method over multiple prototypes, named OpenNCD, is proposed, which consists of two reciprocally enhanced parts. First, a bi-level contrastive learning method is introduced to maintain the pair-wise similarity in both prototype and prototype group levels for better representation learning. The involved prototypes and prototype groups can be regarded as representative points for similar instances on fine-level (sub-classes hierarchy) and coarse-level (real-classes hierarchy). Then, a new and reliable similarity metric is proposed based on the Jaccard distances of common representing instances. The most similar prototypes will be grouped together to represent the same class. Finally, the prototype groups are associated with real labeled classes for known class recognition and novel class discovery.

Our contributions are summarized as follows:
\begin{itemize}
\item We propose a new approach to simultaneously tackle the three challenges in open-world learning.
\item We introduce a bi-level contrastive learning approach to achieve better representation learning. 
\item We design a novel and reliable approach for novel class discovery by progressively grouping the prototypes.
\end{itemize}

\section{Related Works}

\subsection{Semi-supervised Learning}

Traditional semi-supervised learning methods \cite{lee2013pseudo,xie2020unsupervised,berthelot2019mixmatch,sohn2020fixmatch,li2021comatch} assume a close-set scenario, in which the classes of unlabeled samples are the same as the labeled samples. However, in the realistic open world, unknown novel classes exist and are often mixed in the unlabeled set, which is very likely to bring significant performance degradation to the classification of labeled known classes. To deal with this problem, some robust open-set semi-supervised learning methods \cite{chen2020semi,yu2020multi,guo2020DS3L,huang2021trash} are proposed. These methods usually assign the detected novel classes as out-of-distribution samples with low weight to decrease the impacts in the training phase, but they cannot identify different novel classes in the detected novel samples. Recently, ORCA \cite{orca} proposes a new approach to further cluster the novel classes while recognizing the known classes. However, it assumes that the number of unknown novel classes is predefined, and that is just impossible for open-world scenarios.

\subsection{Novel Class Discovery}
Novel class discovery tasks aim to cluster the unlabeled novel classes with the help of a similar but disjointed extra labeled set, as proposed in several existing approaches \cite{han2019learning,han2019automatically,brbic2020mars,zhong2021openmix}. However, it is assumed that all samples belong to the novel unknown classes in the testing phase. Therefore, while these approaches are able to discover novel classes in the unlabeled set, they cannot identify the known classes which surely exist in the open-world setting.
Recently, GCD \cite{gcd} generalizes the novel class discovery task to further recognize the known classes in the unlabeled set. It relies on a well-train vision transformer (ViT) model \cite{dosovitskiy2020vit} to extract better visual representation. However, in the testing phase, all testing samples are required for clustering to get the final predictions, which is unable to make predictions directly.

\subsection{Class Number Estimation}
In the open-world scenario, the number of classes or clusters can not be obtained in advance. 
To estimate the number of classes during the procedure of clustering, most traditional methods \cite{pelleg2000x-means,hamerly2003G-means,kalogeratos2012dip-means} first initialize a certain number of clusters, and then apply a criterion to determine whether the clusters should be split or merged in an iteration way. In the field of deep clustering, only a few recent works \cite{leiber2021dip,ronen2022deepdpm} include the approach for cluster number estimation. However, the samples in the dataset can be clustered under different hierarchy levels since one class category can be further divided into several sub-categories. Therefore, it is difficult to choose the thresholds of criterion to get the most suitable cluster hierarchy. In the open-world scenario, one can determine the cluster hierarchy in the unlabeled set by matching it with the labeled data to better estimate the number of classes.

\begin{figure*}[htp]
\centering 
\includegraphics[width = 0.84\textwidth]{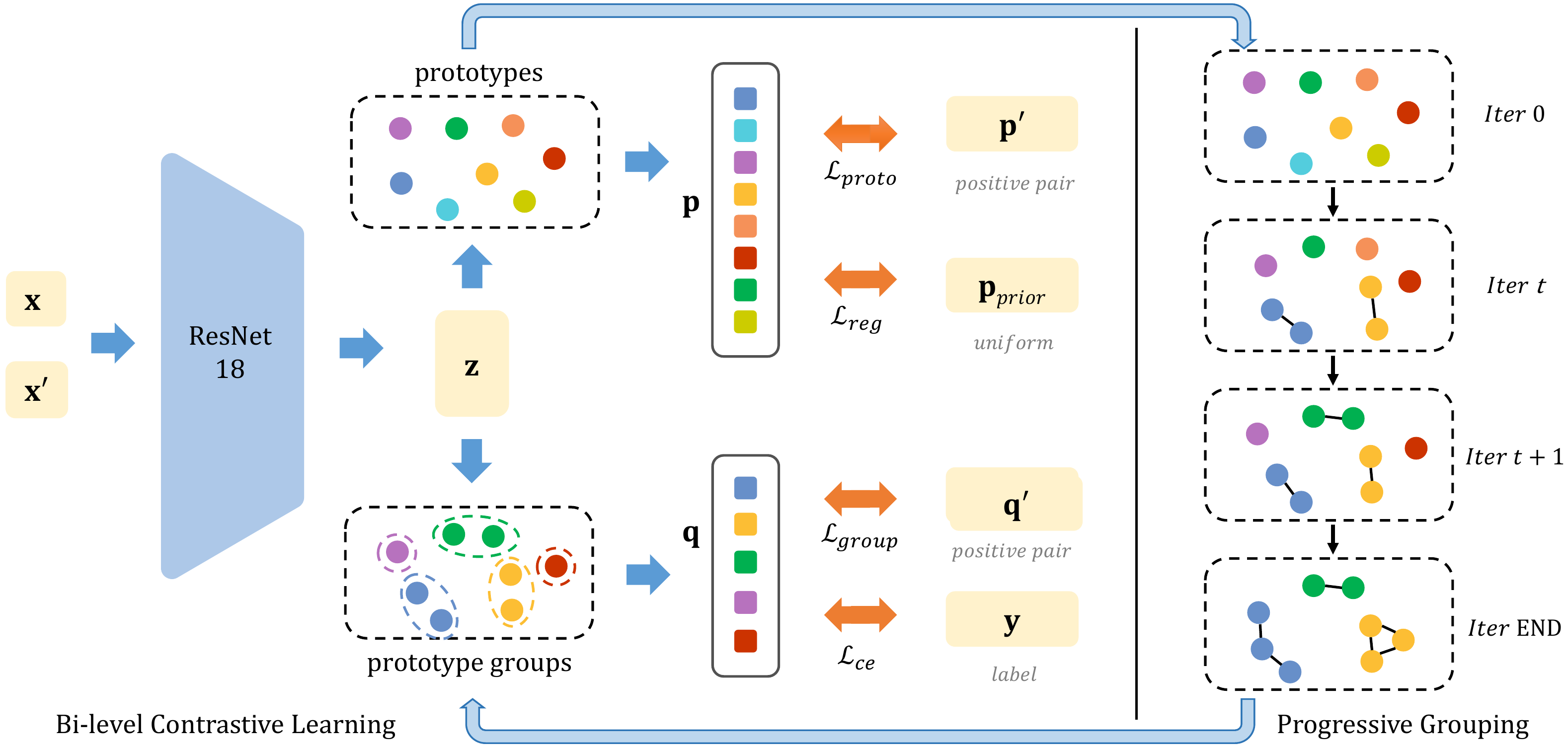}
\caption{The proposed OpenNCD consists of two reciprocally enhanced parts.
First, the bi-level contrastive learning method maintains the similarity in positive pairs $x$ and $x'$ on prototype level $\mathcal{L}_{proto}$ and prototype group level $\mathcal{L}_{group}$ for better representation learning. The prototype prior regularization term $\mathcal{L}_{reg}$ prevents the prototypes from collapsing, while the multi-prototype cross-entropy term $\mathcal{L}_{ce}$ aims to maintain the known class recognition performance. Similar prototypes are grouped progressively as the model's representing ability improves.}
\label{fig:framework}
\end{figure*}

\section{Proposed Method}

\subsection{Problem Definition}
Given a dataset consisting of the labeled part $\mathcal{D}_l$ and unlabeled part $\mathcal{D}_u$. We denote $\mathcal{D}_l = \{(x_{i},y_{i})\}_{i=1}^{m}$, where label $y_{l}$ is from the set of classes $\mathcal{Y}_l$. We also denote $\mathcal{D}_u = \{(x_i)\}_{i=1}^n$, where the ground-truth label $y_{u}$ is from the unknown set of classes $\mathcal{Y}_u$. In our open-world setting, the unlabeled set $\mathcal{D}_u$ consists of several novel classes that do not exist in the labeled set, namely, $\mathcal{Y}_l \subset \mathcal{Y}_u$. We consider $\mathcal{Y}_{novel} = \mathcal{Y}_u/\mathcal{Y}_l$ as the set of novel classes. In addition, different from the existing methods that obtain the number of novel classes in advance, in our setting, the number of novel classes $\left |\mathcal{Y}_{novel}\right |$ is an unknown value that requires estimation.

\subsection{Framework}

To handle this challenging task, we introduce a novel progressive bi-level contrastive learning method over multiple prototypes. The multiple prototypes are trained to represent the instances in the feature space. A bi-level semi-supervised contrastive learning approach is designed to learn better representations. A new approach is also proposed to group similar prototypes in a progressive way for unknown novel class discovery.

As shown in Figure \ref{fig:framework}, the whole framework is composed of a feature extractor $f_\theta$ and multiple trainable prototypes $\mathcal{C} = \{\mathbf{c}_{1}, . . . , \mathbf{c}_{K}\}$ in the feature space. The number of prototypes $K$ is predefined, which far exceeds the number of potential real classes. We also conduct an experiment with the effect of different $K$. The assignment probabilities from the instances to the prototypes are calculated and further utilized for the bi-level semi-supervised contrastive learning method to learn better representations for class discovery.

To group the unlabeled prototypes and discover the classes, a new metric based on common representing instances is proposed to measure the similarity of two prototypes that belong to the same class. The prototypes are grouped progressively at the training stage.
As the representation ability of the encoder networks increases in the iterative training, the instance from the same class will be more compact and the associated prototypes will also have a higher similarity. In this way, we can find reliable groups and estimate the number of novel classes.

The objective function of the proposed approach includes four components:
\begin{equation}
\mathcal{L} = \mathcal{L}_{proto} + \mathcal{L}_{group} + \lambda_{1} \mathcal{L}_{reg} + \lambda_{2} \mathcal{L}_{ce} 
\end{equation}
where the first two terms are the bi-level contrastive loss, consisting of the  prototype-level similarity loss $\mathcal{L}_{proto}$ and the group-level similarity loss $\mathcal{L}_{group}$. $\mathcal{L}_{reg}$ is the prototype entropy regularization loss and $\mathcal{L}_{ce}$ is the multi-prototype cross-entropy loss.

\subsection{Bi-level Contrastive Learning}

\paragraph{Prototype-level Similarity.}
Given an instance feature $\mathbf{z}$ from the feature extractor and a set of prototypes $\mathcal{C} = \{\mathbf{c}_{1}, . . . , \mathbf{c}_{K}\}$, we denote $\mathbf{p}\in \mathbb{R}^{(m+n)\times K}$ as the assignment probability from $\mathbf{z}$ to the prototype set $\mathcal{C}$ based on the cosine similarity. And the assignment probability from $\mathbf{z}$ to the $k$th prototype $\mathbf{c}_{k}$ is given by:
\begin{equation}
    \mathbf{p}^{(k)}
  = \frac{ \exp \left ( \frac{1}{\tau} \mathbf{z} \cdot \mathbf{c}^\top_k \right ) }
  {\sum_{\mathbf{c}_{k'}\in \mathcal{C} } \exp \left ( \frac{1}{\tau} \mathbf{z} \cdot\mathbf{c}^\top_{k'} \right ) },
    \label{eq:q}
\end{equation}
where $\tau$ is the temperature scale, and both $\mathbf{z}$ and $\mathbf{c}$ are $l_{2}$ normalized.

The pair-wise semi-supervised contrastive learning technique is used for representation learning. Each instance in a training batch is treated as an anchor point. The positive point is obtained either by selecting another random instance from the same class if it is labeled or by selecting its nearest neighbor instance if it is unlabeled. Then the augmented view of the anchor point is chosen to calculate the similarity with the selected positive point to create more discrepancy for contrastive learning.

To make the two instances in a positive pair exhibit similar assignment probabilities, we define the prototype-level similarity loss as:
\begin{equation}
  \mathcal{L}_{proto} = - \frac{1}{m+n}  \sum_{i}^{m+n}   
  \log \langle \mathbf{p}_{i}, \mathbf{p}_{i}' \rangle,
  \label{eq:protosim}
\end{equation}
where ${p}_{i}$ and ${p}_{i}'$ denote the assignment probability of the anchor point and its positive point, respectively. $\langle \cdot \rangle$ represents the cosine distance.  
Note that we directly utilize the instances in the same training mini-batch as positive samples to update the encoder and prototypes, instead of sampling from the entire dataset for negative samples. In this way, the proposed model can be trained and updated online for large-scale datasets.

\paragraph{Group-Level Similarity.}
The above prototype-level similarity loss makes the instances represented by the fine-grained prototypes. To learn better representation for class identification, the instances should also be represented on the more coarse-grained level to match the real class hierarchy. To this end, we introduce a group-level similarity loss by representing the instances on the level of prototype groups.

Formally, a prototype group $\mathcal{C}_{g}$ is a set of similar prototypes that are very likely to represent the same class, where $\mathcal{C}_{g} \subset \mathcal{C}$ and any two prototype groups are disjoint. The prototypes are grouped in a progressive way for class discovery, and the detailed grouping process is illustrated in Section \ref{prototype grouping}.

Then, the assignment probability from instance feature $\mathbf{z}$ to prototype group $\mathcal{C}_{g}$ can be obtained by: 
\begin{equation}
  \mathbf{q}_{i}^{(g)}
  = \frac{\sum_{\mathbf{c}_{k}\in \mathcal{C}_{g} } \exp \left ( \frac{1}{\tau} \mathbf{z} \cdot \mathbf{c}^\top_k \right ) }
  {\sum_{\mathbf{c}_{k'}\in \mathcal{C} } \exp \left ( \frac{1}{\tau} \mathbf{z} \cdot\mathbf{c}^\top_{k'} \right ) }.
  \label{eq:p}
\end{equation}

And the group-level similarity loss can be denoted as:
\begin{equation}
  \mathcal{L}_{group} = 
  - \frac{1}{m+n}  \sum_{i}^{m+n}   
(\mathbf{q}_{i}'\log  \mathbf{q}_{i}  + \mathbf{q}_{i}\log  \mathbf{q}_{i}'), 
  \label{eq:groupsim}
\end{equation}
where ${q}_{i}$ and ${q}_{i}'$ denote the assignment probability of two instances in a positive pair, which can be regarded as the pseudo label for each other on the group level reciprocally. Note that here we utilize another form of contrastive similarity loss as Equation \ref{eq:protosim} to extract knowledge from a different perspective to prevent overfitting. Moreover, Equation \ref{eq:groupsim} has the same form as cross-entropy, which makes the model focus more on group or class discrimination.

\paragraph{Prototype Regularization.}
Since multiple prototypes are utilized to represent the feature distribution, some prototypes or groups with low assignment probabilities might be ignored during the optimization. More seriously, all of the instances are assigned to the same prototype in some cases, leading to model collapse. To solve this problem, we introduce a prototype regularization term to regularize the marginal assignment probability on prototypes $\mathbf{p}_{proto} \in \mathbb{R}^{K}$ to be close to a balance prior $\mathbf{p}_{prior}$ by Kullback-Leibler (KL) divergence, which is given by:
\begin{equation}
  \mathcal{L}_{reg} = KL \left (\mathbf{p}_{proto} \ \| \  \mathbf{p}_{prior} \right ),
  \label{eq:ent_kl}
\end{equation}
where
\begin{equation}
    \mathbf{p}_{proto} = \frac{1}{m+n}  \sum_{i}^{m+n}   \mathbf{p}_{i},
    \label{eq:p_proto}
\end{equation}

We expect to prevent the possible collapsing when all instances are assigned to the same single prototype group, and meanwhile, ensure that all the prototypes are utilized to represent distinct characteristics of the complex class distribution. Therefore, we design the prior $\mathbf{p}_{prior}$ to bring a uniform distribution among the prototype groups and also among the prototypes in one group, which is denoted as:
\begin{equation}
    \mathbf{p}_{prior}^{(k)} = \frac{1}{\mathcal{N}_{g}\times \left | \mathcal{C}_{k} \right |},
    \label{eq:p_prior}
\end{equation}
where $k$ denotes the prior of the $k$th prototype $\mathbf{c}_{k}$, $\mathcal{N}_{g}$ is the number of all prototype groups in the current stage, and $ \left | \mathcal{C}_{k} \right |$ is the number of prototypes of the group which $\mathbf{c}_{k}$ belongs to.

\paragraph{Multi-Prototype Cross Entropy.}
The above objectives mainly focus on unsupervised representation learning. To further improve the capability of known class recognition, the supervised multi-prototype cross-entropy loss is introduced.

Note that the prototype groups are unlabeled in the progressive grouping stage, we hope that the prototype groups can be matched to the real class levels where each prototype group represents one class. To this end, we utilize the Hungarian algorithm \cite{hw1955hungarian} to match the known classes with the prototype groups. In this way, for the labeled instances, we obtain their ground truth label on the prototype groups. Then, the supervised multi-prototype cross-entropy loss is defined as:
\begin{equation}
  \mathcal{L}_{ce} = - \frac{1}{m} \sum_{i}^{m} (\log  \mathbf{q}_{i}^{(y)}), 
  \label{eq:cls}
\end{equation}
where $\mathbf{q}_{i}^{(y)}$ is the assignment probability on the ground-truth group, which is given by:
\begin{equation}
\mathbf{q}_{i}^{(y)}
  = \frac{\sum_{\mathbf{c}_{k}\in \mathcal{C}_{y} } \exp \left ( \frac{1}{\tau} \mathbf{z}_{i} \cdot \mathbf{c}^\top_k\right ) }
  {\sum_{\mathbf{c}_{k'}\in \mathcal{C}} \exp \left ( \frac{1}{\tau} \mathbf{z}_{i} \cdot\mathbf{c}^\top_{k'} \right ) }.
  \label{eq:cls_where}
\end{equation}

Note that Equation \ref{eq:cls} has a similar form as Equation \ref{eq:groupsim}. The difference is that Equation \ref{eq:groupsim} regards the group assignment probability of the other view as the pseudo label while Equation \ref{eq:cls} utilizes the ground-truth labels directly.

\begin{figure}[htp]
\centering 
\includegraphics[width = 0.43\textwidth]{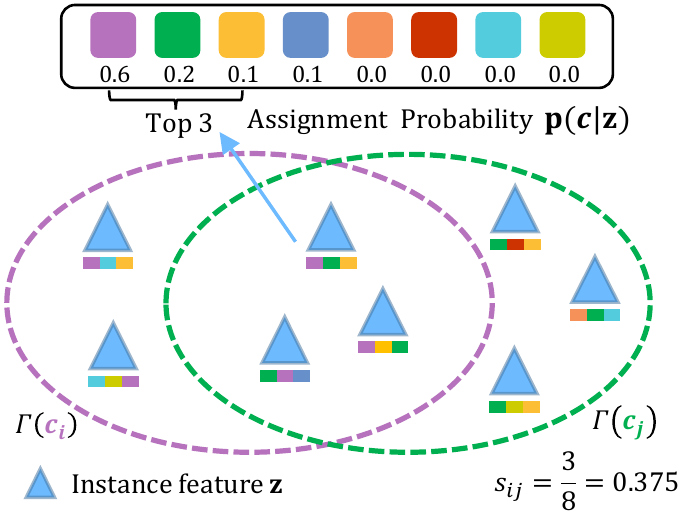}
\caption{Similarity metric. In this example, each instance has 3 associated prototypes (with top 3 assignment probabilities over all prototypes). The similarity score is computed by Jaccard distance over the representing instance sets of two prototypes.}
\label{fig:grouping}
\end{figure}

\subsection{From Prototypes to Classes: Progressive Prototype Grouping}
\label{prototype grouping}
In this section, we elaborate on the process of prototype grouping with the aim of progressively discovering the unknown number of novel classes.

First, a novel similarity metric is proposed to judge the similarity between two prototypes, which is shown in Figure \ref{fig:grouping}. Specifically, for an instance feature $\mathbf{z}$, its assignment probability $\mathbf{p}$ is sorted on all prototypes. The prototypes with top $\kappa$ largest assignment probabilities are regarded as the associated prototypes of $\mathbf{z}$, and in turn, $\mathbf{z}$ is regarded as a representing instance of prototype $\mathbf{c}$. The set of all representing instances of prototype $\mathbf{c}$ is denoted as:

\begin{equation}
    \Gamma ( \mathbf{c} )  = \left \{ \mathbf{z}_i |(\mathbf{c} \in \mathrm{Top\kappa }(\mathbf{q}_i)) \right \}.
\end{equation}

Intuitively, two prototypes are more likely to belong to the same class if they have more common representing instances. Therefore, we calculate the similarity score of two prototypes by the Jaccard distance over their representing instance sets, given by:
\begin{equation}
    s_{ij} = \frac{\left | \Gamma (\mathbf{c}_{i}) \cap \Gamma (\mathbf{c}_{j})  \right | }{\left | \Gamma (\mathbf{c}_{i}) \cup \Gamma (\mathbf{c}_{j})  \right |}. 
\end{equation}

At each epoch, the similarity between each two prototypes is calculated to form an affinity matrix. Some graph-based clustering methods can then be utilized to detect the densely connected prototypes, which are regarded as prototype groups. A simple way is to find the prototype groups by linked prototypes, where the similarities are over a certain threshold $\delta$. Some approaches that do not require a predefined number of classes can also work, such as Louvain \cite{blondel2008fast} and affinity propagation \cite{frey2007clustering}. To ensure the novel classes are clustered into the same hierarchical level as the known classes, the threshold $\delta$ (or the parameter to control the clustering hierarchy) is adjusted automatically by achieving the highest accuracy on the labeled known class samples. With the increasing representation ability of the feature extractor and prototypes, more reliable prototype groups can be obtained gradually.

\begin{table*}[ht]
	\centering
	\setlength{\tabcolsep}{3mm}{
	\begin{tabular}{lcccccccccccccc}
		\toprule
		 & \multicolumn{3}{c}{\textbf{CIFAR-10}}  &                    
		 & \multicolumn{3}{c}{\textbf{CIFAR-100}} &
		 & \multicolumn{3}{c}{\textbf{ImageNet-100}}\\ 
		\textbf{Methods}         & \multicolumn{1}{c}{\textbf{Known}} & \multicolumn{1}{c}{\textbf{Novel}} &
		\multicolumn{1}{c}{\textbf{All}}  & & \multicolumn{1}{c}{\textbf{Known}} & \multicolumn{1}{c}{\textbf{Novel}} &
		 \multicolumn{1}{c}{\textbf{All}}  & &
		 \multicolumn{1}{c}{\textbf{Known}} & \multicolumn{1}{c}{\textbf{Novel}} &
		 \multicolumn{1}{c}{\textbf{All}}\\
		\midrule
		
		{FixMatch}$^\dagger$ & 64.3 & \, 49.4$^\dagger$ & 47.3 & \quad & 30.9 & \, 18.5$^\dagger$ & 15.3 & \quad  & 60.9 & \, 33.7$^\dagger$ & 30.2  \\
		{$\text{DS}^3 \text{L}$}$^\dagger$ & 70.5 & \, 46.6$^\dagger$ & 43.5 & \quad & 33.7 & \, 15.8$^\dagger$ & 15.1 & \quad & 64.3 & \, 28.1$^\dagger$ & 25.9 \\
		{DTC}$^*$ &  \, 42.7* & 31.8 & 32.4 & \quad & \, 22.1* & 10.5  & 13.7 & \quad & \, 24.5* & 17.8  & 19.3 \\
		{RankStats}$^*$ &  \, 71.4* & 63.9 & 66.7 & \quad & \, 20.4* & 16.7 & 17.8 & \quad & \, 41.2* & 26.8 & 37.4  \\
		{SimCLR}$^*$ & \, 44.9* & 48.0 & 47.7 & \quad & \, 26.0* & 28.8 & 26.5 & \quad  & \, 42.9* & 41.6 & 41.5  \\
		ORCA &  82.8 & 85.5 & 84.1 & \quad &  52.5 & 31.8 & 38.6 & \quad  &  83.9 & 60.5 & 69.7  \\
		GCD &  78.4 & 79.7 & 79.1 & \quad &  49.7 & 27.6 & 38.0 & \quad  &  82.3 & 58.3 & 68.2  \\
		\midrule
		{\textbf{OpenNCD}} &\textbf{83.5} & \textbf{86.7} & \textbf{85.3} & \quad & \textbf{53.6} & \textbf{33.0} & \textbf{41.2} & \quad & \textbf{84.0} & \textbf{65.8} & \textbf{73.2}   \\
		\bottomrule
	\end{tabular}}
    \caption{Accuracy comparison of known, novel, and all classes. The dataset is composed of 50\% known classes and 50\% novel classes, with only 10\% of known classes labeled. Asterisk ($\ast$) denotes that the original method cannot recognize known classes, which are extended by matching the discovered clusters to the classes in the labeled data. Dagger ($\dagger$) denotes that the original method cannot discover novel classes, which are extended by performing clustering on the out-of-distribution samples.}
    \label{tab:old_new_all_acc}
	\vspace{-1em}
\end{table*}

\section{Experiments}
\subsection{Experimental Setup}

\paragraph{Datasets.}
Our proposed approach\footnote{Code at \url{https://github.com/LiuJMzzZ/OpenNCD}\label{code}} is evaluated on three widely-used datasets in the standard image classification tasks including CIFAR-10 \cite{krizhevsky09learning}, CIFAR-100 \cite{krizhevsky09learning}, and ImageNet-100, which is randomly sub-sampled with 100 classes from ImageNet \cite{deng2009imagenet} for its large volume. For each dataset, the first 50\% classes are regarded as known and the rest 50\% as unknown. Further, only 10\% of the known classes are labeled, with the unknown and the rest of the known all unlabeled.

\paragraph{Evaluation Metrics.}
Following the evaluation protocol in \cite{orca}, we evaluate performance on both the known and the novel classes. For known classes, the classification accuracy is measured in the testing phase. For unknown classes, the clustering accuracy is calculated by solving the prediction-target class assignment based on the Hungarian algorithm \cite{hw1955hungarian}. Moreover, we also calculate the clustering accuracy on all known and novel class samples to measure the overall performance of the proposed model.

\paragraph{Implementation Details.}
In the proposed method, Resnet-18 is used as the backbone of the feature extractor, which is pre-trained by SimCLR \cite{chen2020simclr} in an unsupervised way. To avoid overfitting, we fix the parameters in the first three blocks of the backbone and only finetune the last block. 50 prototypes are utilized for the CIFAR-10 dataset and 500 for both the CIFAR-100 and ImageNet-100 datasets with a fixed dimension of 32. We adopt an Adam optimizer with a learning rate of 0.002 and fix the batch size to 512 in all experiments. The temperature scale $\tau$ is set to 0.1 suggested by most of the previous methods, and the weight of the last two terms in the objective function is set to $\left\{\lambda_{1}, \lambda_{2}\right\}=\left\{1, 1\right\}$. $\kappa$ is set to 5 in prototype grouping.

\subsection{Comparison with Baselines}

\paragraph{Baselines.}
Our proposed method is compared with two baseline methods ORCA \cite{orca} and GCD \cite{gcd}. ORCA is proposed for the open-world semi-supervised learning problem with a predefined number of novel classes. For consistency, our experiment settings are following the same protocol of ORCA from its public code. GCD utilizes a more sophisticated pre-trained backbone which exhibits high performance during initialization, and we replace it in the public code with the Resnet-18 network same as in our settings for a fair comparison. Note that in the original paper of ORCA, the ratio of labeled data of the known classes is set to 50\%, which is a relatively high ratio for semi-supervised learning. Here we consider comparing in a tougher and general scenario, where only 10\% known class data are labeled. We also report the result with the labeled ratio of 50\% as an additional result in Table \ref{tab:old_new_all_acc_ratio_05} in the appendix.

\paragraph{Extended Baselines.}
We also compare the recent novel class discovery methods and robust semi-supervised learning methods by extending them into our open-world settings. Novel class discovery methods can only cluster novel classes but cannot classify the unlabeled samples to the known classes. Here, we compare two novel class discovery methods: DTC \cite{han2019learning} and RankStats \cite{han2019automatically}. To extend them for known class recognition, we regard the known class samples as unknown and detect them the same way as normal unknown classes. Then we report the accuracy by using the Hungarian algorithm for label matching. The traditional semi-supervised learning methods cannot discover novel classes mixed in the unlabeled data. We select two methods: FixMatch \cite{sohn2020fixmatch} and DS3L \cite{guo2020DS3L}, and extend them for novel class discovery. The samples with lower confidence scores or weights are selected as unknown class samples, which are clustered by k-means, and the clustering accuracy is reported.

\paragraph{Comparison Results.}
The results of all compared methods are presented in Table \ref{tab:old_new_all_acc}, including the accuracy of known classes, novel classes, and all classes. As an additional baseline, we also run k-means directly on the output features of the encoder which is pre-trained by SimCLR and include the obtained results without extra training. Since most baselines cannot deal with the class number estimation task, we assume the number of novel classes is known for all methods. For the extended method, the "all" scores are not the average of the "known" and "novel" scores, because the marked scores ($\ast$ or $\dagger$) are calculated in an extended and different way. The results in Table \ref{tab:old_new_all_acc} demonstrate that simultaneously recognizing known classes and clustering novel classes in the open-world setting is difficult for traditional methods. However, our proposed method can handle these complex tasks effectively with better performance than baselines.

\begin{table}[tp]
    \centering
    \setlength{\tabcolsep}{2mm}{
    \begin{tabular}{@{}ccccc@{}}
    \toprule[1pt]
    \textbf{Methods}          & \textbf{PredNum}         & \textbf{Acc}        & \textbf{NMI}          \\ \midrule
    X-means          & 13.8 $\pm$ 6.3               & 38.3 $\pm$ 3.1         & 37.2 $\pm$ 9.3              \\
    G-means         & 28.2 $\pm$ 2.1                & 35.8 $\pm$ 0.1         & 50.8 $\pm$ 0.4           \\
    DipDeck          & 8.16 $\pm$ 0.8              & 66.8   $\pm$ 4.7       & 67.0 $\pm$ 1.3               \\  \midrule
    \textbf{Ours(0\% label)}  & \textbf{9.4  $\pm$ 1.1}              & \textbf{78.3  $\pm$ 1.6}        & \textbf{71.2 $\pm$ 2.1}                \\
    Ours(10\% label) & 9.8  $\pm$ 0.4       &       86.2  $\pm$ 0.5    &     79.6 $\pm$ 0.6                \\ \bottomrule[1pt]
    \end{tabular}}
    \caption{Class number estimation in the unlabelled data. We report the results on the CIFAR-10 dataset averaged by 5 runs.}
    \label{tab:numofclass}
\end{table}

\subsection{Estimating the Number of Classes}
The previous experiments assume that the number of novel classes is known, so we further conduct an experiment under the scenario where the real number of novel classes is unknown. To this end, the proposed method is compared with some traditional class number estimation methods including X-means \cite{pelleg2000x-means}, G-means \cite{hamerly2003G-means}, and a recent deep clustering method DipDeck \cite{leiber2021dip}. For the above three methods, we extract the features of the CIFAR-10 dataset from the pre-trained backbone, which is the same as in our method. Since these three methods are all unsupervised, we report the results of our method in two settings including (a) trained with no labeled data, as in the three methods, and (b) trained with 50\% known and 50\% novel and 10\% of the known labeled, as in the previous experiments. The results in Table \ref{tab:numofclass} demonstrate that our proposed method can better deal with the class number estimation task even without the labeled data.

\begin{table}[tp]
	\centering
	\setlength{\tabcolsep}{5mm}{
	\begin{tabular}{ccccc}
		\toprule
		\textbf{Methods}         & \multicolumn{1}{c}{\textbf{Seen}} & \multicolumn{1}{c}{\textbf{Novel}} &
		\multicolumn{1}{c}{\textbf{All}} \\
		\midrule
		\textbf{w/o  $\mathcal{L}_{reg}$} & 11.9 & 14.2 & 29.3 \\
		\textbf{w/o  $\mathcal{L}_{ce}$} & 27.9 & 25.6 & 23.5 \\
		\textbf{w/o  $\mathcal{L}_{group}$} & 44.0 & 26.0 & 33.5 \\
		\textbf{w/o  $\mathcal{L}_{proto}$} & 50.3 & 31.7 & 39.7  \\
		\textbf{OpenNCD}  & \textbf{53.6} & \textbf{33.0} & \textbf{41.2}  \\
		\bottomrule
	\end{tabular}}
	\caption{Ablation analysis on the components of the objective function. We report the results on the CIFAR-100 dataset with 50\% known and 50\% novel and 10\% of the known labeled.}
    \label{tab:ablation}
\end{table}

\subsection{Ablation and Analysis}

\paragraph{Ablation Study.}
In Table \ref{tab:ablation}, the contributions of different parts of the loss functions in our proposed approach are analyzed, including the prototype-level similarly $\mathcal{L}_{proto}$, the group-level similarity $\mathcal{L}_{group}$, the multi-prototype cross entropy $\mathcal{L}_{ce}$ and the prototype regularization $\mathcal{L}_{reg}$. To investigate the importance of these terms, the ablation study is conducted by removing each term separately from the objective function. We can infer from Table \ref{tab:ablation} that all the components contribute to our proposed method. Moreover, the result with $\mathcal{L}_{group}$ removed demonstrates the importance of grouping, and the result with $\mathcal{L}_{proto}$ removed proves the benefit of multiple prototypes in representation learning.

\begin{figure*}[ht]
    \centering
	\includegraphics[width=0.95\textwidth]{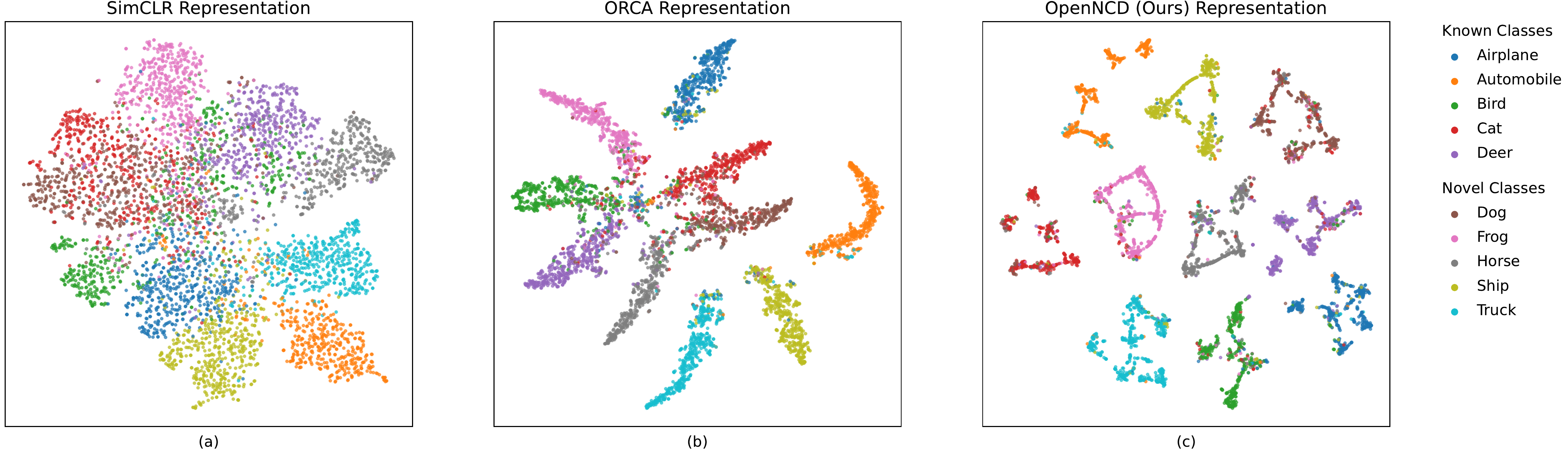}
    \caption{T-SNE visualization of learned feature representation for CIFAR-10 with 50\% known classes (10\% labeled) and 50\% novel classes on (a) pre-trained ResNet-18 by SimCLR, (b) ORCA (Baseline), and (c) OpenNCD (Ours) with 50 prototypes. Colors represent classes.}
    \label{fig:hidden_space}
\end{figure*}

\begin{figure}[tp]
\centering 
\includegraphics[width = 0.38\textwidth]{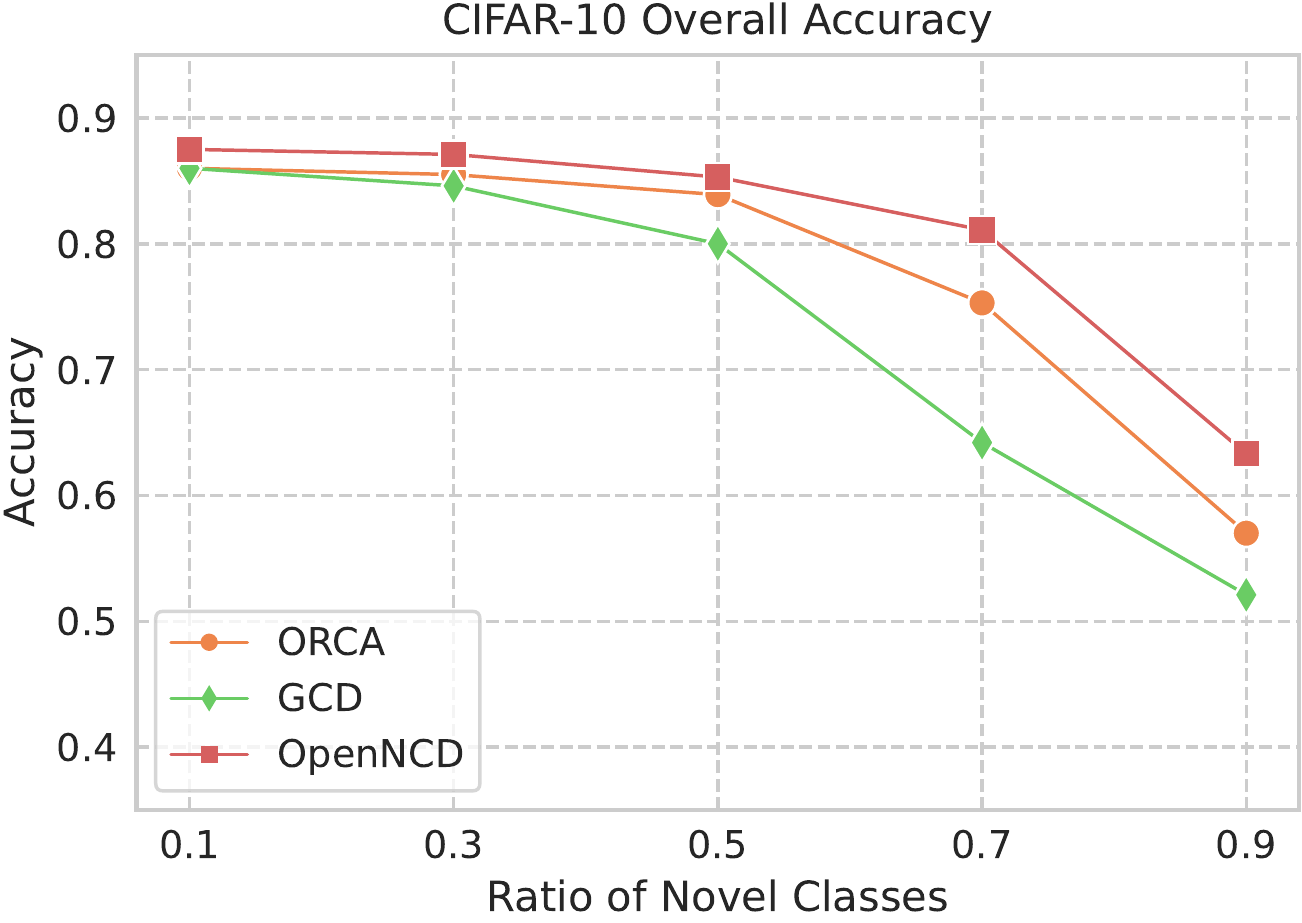}
\caption{Performance of ORCA, GCD, and the proposed method on the CIFAR-10 dataset with different novel class ratios.}
\label{fig:novel-ratio-cifar100}
\end{figure}

\paragraph{Effect of the Novel Class Ratio.}
As shown in Figure \ref{fig:novel-ratio-cifar100}, we evaluate the performance on the CIFAR-10 dataset with the ratio of novel classes ranging from 0.1 to 0.9. We report the accuracy of all classes by unsupervised clustering accuracy, which can be considered as a good proxy for the overall performance. Despite the inevitable performance degradation due to the increase in the novel class ratio, OpenNCD still performs better than the two strong baselines, especially at higher novel class ratios. The result in Figure \ref{fig:novel-ratio-cifar100} indicates that our proposed approach can better face task scenarios with higher openness.

\begin{figure}[htp]
\centering 
\includegraphics[width = 0.38\textwidth]{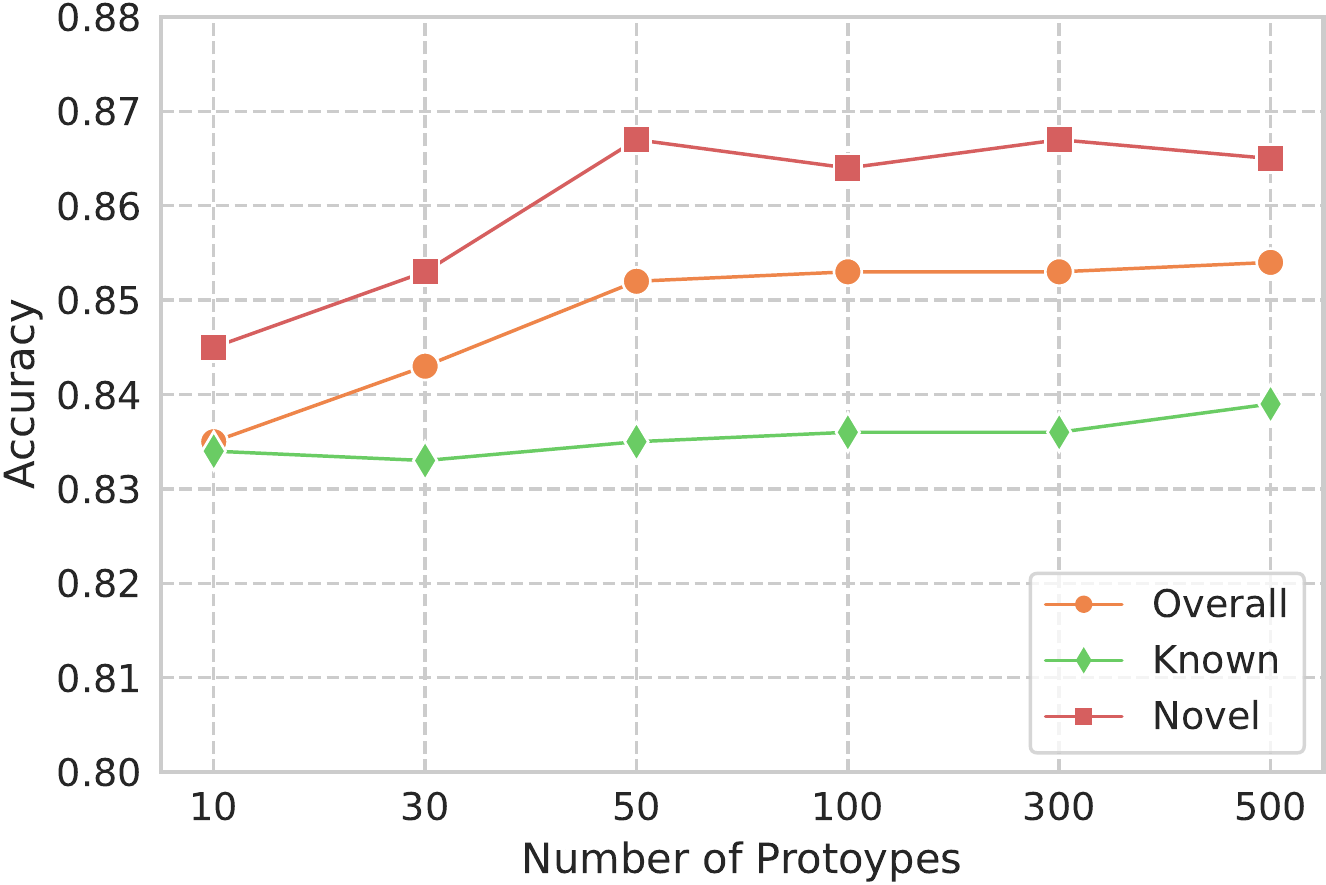}
\caption{Performance on the CIFAR-10 dataset with number of prototypes varying from 10 (single prototype for each class) to 300.}
\label{fig:num-of-proto}
\end{figure}

\paragraph{Benefits of Multiple Prototypes.}
To investigate the effect of the number of prototypes $K$, we conducted experiments by increasing it from the actual number of classes (10) to a large number (500) on the CIFAR-10 dataset. The results are shown in Figure \ref{fig:num-of-proto}. It can be observed that using multiple prototypes rather than a single one, to represent the distribution of each class, improves the performance of both known and novel classes. Moreover, representing class distributions using a single prototype requires prior knowledge of the number of new classes. When the number of new classes is unknown, multiple prototypes are employed to represent the class distribution, which will be grouped progressively to discover novel classes. Further increasing the number of prototypes does not result in continuous improvement in representation learning, and the performance tends to stabilize. We can still infer that more prototypes would be beneficial when the data distribution is more complex and harder to approximate.

\paragraph{Visualization of the Feature Space.}
In Figure \ref{fig:hidden_space}, we show the learned latent space of the raw pre-trained ResNet-18, ORCA, and the proposed OpenNCD on the CIFAR-10 dataset. The high-dimensional latent features are reduced to 2D by T-SNE \cite{van2008visualizing}. As shown in Figure \ref{fig:hidden_space}(a), the raw ResNet-18 features are mixed with each other and do not form clear clusters. In the latent space of ORCA shown in Figure \ref{fig:hidden_space}(b), the instance features are more separated, but they are of irregular shapes and do not cluster compactly enough. Comparatively, Figure \ref{fig:hidden_space}(c) shows the latent space of the proposed OpenNCD, where the features are represented by 50 prototypes and 10 prototype groups represented by different colors. We can see the prototypes are linked in each group and features distribute closely around the associated prototypes and groups, forming very compact and clear clusters, which benefits our progressive prototype grouping approach for discovering novel classes.

\section{Conclusion}
In this paper, we tackle the three challenges simultaneously in the open-world setting, including known class recognition, novel class discovery, and class number estimation. To this end, we propose a novel method named OpenNCD with two reciprocally enhanced parts. First, a bi-level contrastive learning method maintains the pair-wise similarity of the prototypes and the prototype group levels for better representation learning. Then, a progressive prototype grouping method based on a novel reliable similarity metric groups the prototypes, which are further associated with real labeled classes for novel class discovery. We conduct extensive experiments and the results demonstrate that our proposed method can deal with the challenges effectively in the open-world setting and outperform the previous methods.



\section*{Acknowledgments}
This work is supported by the Fundamental Research Funds for the Central Universities (ZYGX2019Z014), Sichuan Key Research Program (22ZDYF3388), National Natural Science Foundation of China (61976044, 52079026), and Fok YingTong Education Foundation for Young Teachers in the Higher Education Institutions of China (161062).


\bibliographystyle{named}
\bibliography{OpenNCD}

\clearpage
\appendix
\section*{\LARGE Open-world Semi-supervised Novel Class Discovery: Appendix}

\vspace{1em}

\section{Training Algorithm}
The training steps of OpenNCD are provided in Algorithm \ref{alg}. The proposed method is composed of two reciprocally enhanced parts, bi-level contrastive learning and progressive prototype grouping. In each epoch, the parameters of the feature extractor $f_\theta$ and the prototypes $\mathcal{C}$ are updated. Then, the new prototype groups are obtained based on the updated $f_\theta$ and $\mathcal{C}$. Finally, the number of the final prototype groups can be regarded as the estimated number of classes.

\begin{algorithm}
\renewcommand{\algorithmicrequire}{\textbf{Input:}}
\renewcommand{\algorithmicensure}{\textbf{Output:}}
\caption{OpenNCD Training Procedure} 
\label{alg}
\begin{algorithmic}[1]
\REQUIRE{Labeled data $\mathcal{D}_l$, unlabeled data $\mathcal{D}_u$, number of prototypes $K$}
\ENSURE{Learned feature extractor $f_\theta$, prototypes $\mathcal{C}$, prototype groups $\mathcal{C}_{g}$ and $EstimatedClassNum$}
\STATE Pretrain $f_\theta$, initialize $\mathcal{C}$ and $\mathcal{C}_{g}$
\WHILE {$epoch < epoch^{max}$}
\WHILE {$batch < batch^{max}$}
        \STATE  $ \mathbf{x} \leftarrow \text{SampleMiniBatch}(\mathcal{D}_l \cup \mathcal{D}_u)$
    	\STATE  $ \mathbf{z} \leftarrow \text{Forward}(\mathbf{x}; f_{\theta})$  
        \STATE  $ \mathbf{p}\leftarrow \text{PrototypeAssignmentProb}(\mathbf{z}, \mathcal{C})$ 
        \STATE  $ \mathbf{q}\leftarrow \text{GroupAssignmentProb}(\mathbf{z}, \mathcal{C}_{g})$
        \STATE $ \mathcal{L}  \leftarrow  \mathcal{L}_{proto} + \mathcal{L}_{group} + \lambda_1 \mathcal{L}_{reg} + \lambda_2 \mathcal{L}_{ce}$
        \STATE $f_\theta, \mathcal{C}  \leftarrow \text{AdamOptimizer}(\mathcal{L})$

\ENDWHILE
\STATE $ \mathcal{C}_{g} \leftarrow \text{PrototypeGrouping(Section 3.4)}$
\ENDWHILE
\STATE $ Estimated ClassNum \leftarrow |\mathcal{C}_{g}|$
\end{algorithmic}
\end{algorithm}
\vspace{-0.5em}

\begin{table*}[bp]

	\centering
	\setlength{\tabcolsep}{3mm}{
	\begin{tabular}{lcccccccccccccc}
		\toprule
		 & \multicolumn{3}{c}{\textbf{CIFAR-10}}  &                    
		 & \multicolumn{3}{c}{\textbf{CIFAR-100}} &
		 & \multicolumn{3}{c}{\textbf{ImageNet-100}}\\ 
		\textbf{Methods}         & \multicolumn{1}{c}{\textbf{Known}} & \multicolumn{1}{c}{\textbf{Novel}} &
		\multicolumn{1}{c}{\textbf{All}}  & & \multicolumn{1}{c}{\textbf{Known}} & \multicolumn{1}{c}{\textbf{Novel}} &
		 \multicolumn{1}{c}{\textbf{All}}  & &
		 \multicolumn{1}{c}{\textbf{Known}} & \multicolumn{1}{c}{\textbf{Novel}} &
		 \multicolumn{1}{c}{\textbf{All}}\\
		\midrule
        {FixMatch}$^\dagger$ & 71.5 & \, 50.4$^\dagger$ & 49.5 & \quad & 39.6 & \, 23.5$^\dagger$ & 20.3 & \quad  & 65.8 & \, 36.7$^\dagger$ & 34.9 \\
		{$\text{DS}^3 \text{L}$}$^\dagger$ & 77.6 & \, 45.3$^\dagger$ & 40.2 & \quad & 55.1 & \, 23.7$^\dagger$ & 24.0 & \quad & 71.2 & \, 32.5$^\dagger$ & 30.8 \\
		{DTC}$^*$ & \, 53.9* & 39.5 & 38.3 & \quad & \, 31.3* & 22.9  & 18.3 & \quad & \, 25.6* & 20.8 & 21.3 \\
		{RankStats}$^*$ & \, 86.6* & 81.0 & 82.9 & \quad & \, 36.4* & 28.4 & 23.1 & \quad & \, 47.3* & 28.7 & 40.3 \\
		SimCLR &  58.3 & 63.4 & 51.7 & \quad &  28.6 & 21.1 & 22.3 & \quad  &  39.5 & 35.7 & 36.9 \\
		ORCA & 88.2 & 90.4 & 89.7 & \quad & 66.9 & 43.0 & 48.1 & \quad & 89.1 & 72.1 & 77.8 \\
		GCD &  78.4 & 79.7 & 79.1 & \quad &  68.5 & 33.5 & 45.2 & \quad  &  82.3 & 58.3 & 68.2  \\
		\midrule
		{\textbf{OpenNCD}} &\textbf{88.4} & \textbf{90.6} & \textbf{90.1} & \quad & \textbf{69.7} & \textbf{43.4} & \textbf{49.3} & \quad & \textbf{90.0} & \textbf{77.5} & \textbf{81.6}   \\
		\bottomrule
	\end{tabular}}
    \caption{Accuracy comparison of known, novel, and all classes. The dataset is composed of 50\% known classes and 50\% novel classes, with 50\% of the known classes labeled. Asterisk ($\ast$) denotes that the original method cannot recognize known classes, which are extended by matching the discovered clusters to the classes in the labeled data. Dagger ($\dagger$) denotes that the original method cannot discover novel classes, which are extended by performing clustering on the out-of-distribution samples.}
	\label{tab:old_new_all_acc_ratio_05}
	\vspace{-2em}
\end{table*}

\section{Experiments on 50\% Labeled Data}
In Section 4.2 of the main paper, experiments are conducted on the dataset with 50\% known classes and 50\% novel classes, and 10\% of the known class samples are labeled. Here the labeled ratio of the known classes is set to 50\%, and the corresponding results are presented in Table \ref{tab:old_new_all_acc_ratio_05}. It can be observed that our proposed method also outperforms the baseline methods with better performances.

\section{Hyper-parameters Analysis}
\begin{figure}[htp]
\centering 
\includegraphics[width = 0.47\textwidth]{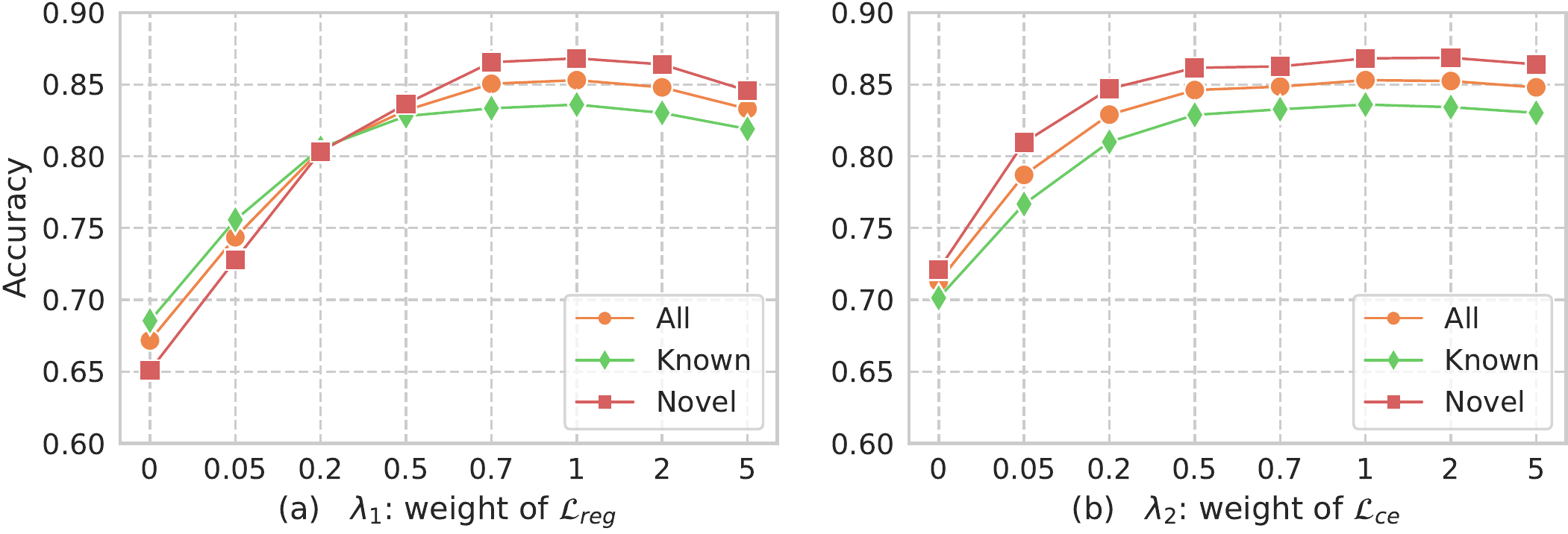}
\caption{Parameter sensitivity analysis on the CIFAR-10 dataset with $\lambda_{1}$ and $\lambda_{2}$ varying from 0 to 5.}
\label{fig:lambda}
\end{figure}

Here, a hyper-parameters sensitivity analysis is conducted on $\lambda_{1}$ and $\lambda_{2}$, which are the weights of the prototype regularization term $\mathcal{L}_{reg}$ and the multi-prototype cross entropy term $\mathcal{L}_{ce}$, respectively. The performances on CIFAR-10 dataset are reported in Figure \ref{fig:lambda} with $\lambda_{1}$ and $\lambda_{2}$ varying from 0 to 5. The results show that both $\mathcal{L}_{reg}$ and $\mathcal{L}_{ce}$ are crucial to the proposed method but remain robust in a wide range of settings.

\vspace{-1em}
\section{Progressive Grouping}
\begin{figure}[htp]
\centering 
\includegraphics[width = 0.35\textwidth]{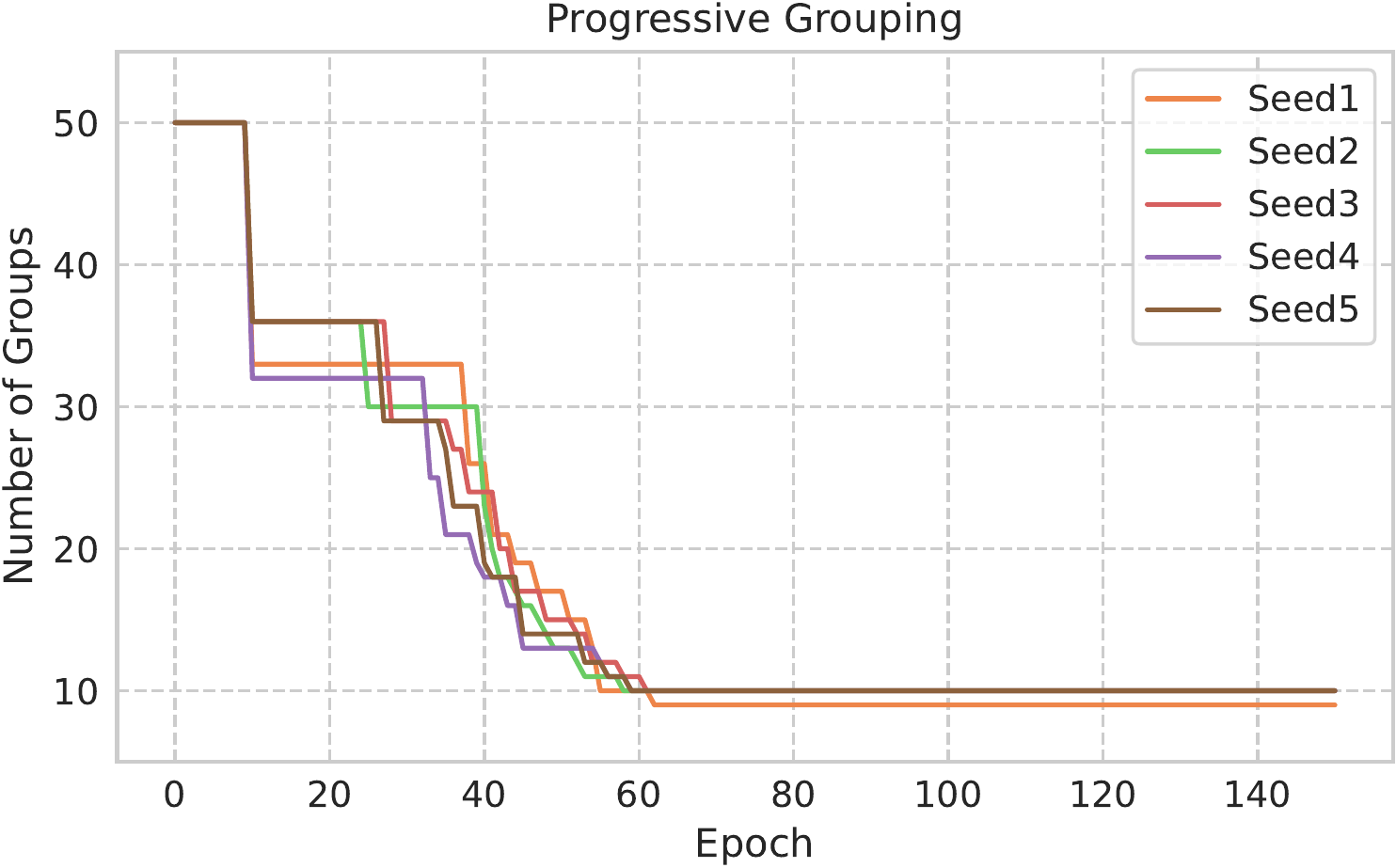}
\caption{Evolution of the number of prototype groups in different epochs. Different colors represent individual runs with different random seeds.}
\label{fig:ProgressiveGrouping}
\end{figure}

We illustrate the progressive grouping procedure by visualizing the evolution of the number of prototype groups across epochs. As shown in Figure \ref{fig:ProgressiveGrouping}, different colors represent individual runs with different random seeds. It can be observed that the number of prototype groups gradually decreases and eventually converges to the actual number of classes.

\end{document}